\documentclass[letterpaper, 10 pt, conference]{ieeeconf}  

\IEEEoverridecommandlockouts                              
\usepackage[utf8]{inputenc}
\usepackage[T1]{fontenc}
\usepackage{graphicx}
\usepackage{array}
\usepackage{tabularx}
\usepackage{booktabs} 
\usepackage{caption} 
\usepackage{url} 

\usepackage{mathptmx} 
\usepackage{amsmath} 
\usepackage{amssymb}  

\title{\LARGE \bf
Sleep Brain and Cardiac Activity Predict Cognitive Flexibility and Conceptual Reasoning Using Deep Learning*
}

\author{Boshra Khajehpiri$^{1}$, Eric Granger$^{1}$, Massimiliano de Zambotti$^{2}$, Fiona C. Baker$^{3}$, Mohamad Forouzanfar$^{1}$
\thanks{© 2025 IEEE. This work was accepted for publication in IEEE EMBC. Personal use of this material is permitted. Permission from IEEE must be obtained for all other uses, in any current or future media, including reprinting/republishing this material for advertising or promotional purposes, creating new collective works, for resale or redistribution to servers or lists, or reuse of any copyrighted component of this work in other works.}
\thanks{*This work was supported by the Natural Sciences and Engineering Research Council of Canada (NSERC) under Grant RGPIN-2021-03924 (MF).}
\thanks{The experimental procedures involving human subjects described in this paper were approved by ÉTS Research Ethics Committee (H20230702).}
\thanks{$^{1}$B. Khajehpiri, E. Granger, and M. Forouzanfar are with the Laboratoire d’imagerie, de vision et d’intelligence artificielle (LIVIA), École de technologie supérieure (ÉTS), Université du Québec, Montreal, QC, Canada}%
\thanks{$^{2}$M. de Zambotti is with Ouraring Inc, San Francisco, CA, USA}%
\thanks{$^{3}$F. C. Baker is with the Center for Health Sciences, SRI International, Menlo Park, CA, USA}%
}

\begin{document}

\maketitle
\thispagestyle{empty}
\pagestyle{empty}

\begin{abstract}

Despite extensive research on the relationship between sleep and cognition, the connection between sleep microstructure and human performance across specific cognitive domains remains underexplored. This study investigates whether deep learning models can predict executive functions, particularly cognitive adaptability and conceptual reasoning from physiological processes during a night’s sleep. To address this, we introduce CogPSGFormer, a multi-scale convolutional-transformer model designed to process multi-modal polysomnographic data. This model integrates one-channel ECG and EEG signals along with extracted features, including EEG power bands and heart rate variability parameters, to capture complementary information across modalities. A thorough evaluation of the CogPSGFormer architecture was conducted to optimize the processing of extended sleep signals and identify the most effective configuration. The proposed framework was evaluated on 817 individuals from the STAGES dataset using cross-validation. The model achieved 80.3\% accuracy in classifying individuals into low vs. high cognitive performance groups on unseen data based on Penn Conditional Exclusion Test (PCET) scores. These findings highlight the effectiveness of our multi-scale feature extraction and multi-modal learning approach in leveraging sleep-derived signals for cognitive performance prediction. To facilitate reproducibility, our code is publicly accessible\footnote{\url{https://github.com/boshrakh95/CogPSGFormer.git}}.

\end{abstract}

\section{INTRODUCTION}

Cognitive decline linked to changes in sleep characteristics—such as variations in sleep architecture, quality, and duration—represents a significant global health challenge. This issue is compounded by the widespread prevalence of diverse sleep patterns and disruptions across populations. Over the past decade, significant connections have been uncovered between sleep patterns and cognitive performance \cite{c1}. Sleep disturbances have been found to have adverse effects on various cognitive abilities, including short-term memory, focused attention, and higher-level cognitive functions like decision-making and problem-solving \cite{c2}. Multiple studies have investigated the relationship between different cognition aspects and subjective and objective sleep measures, including self-reported questionnaires, actigraphy, and polysomnography (PSG) signals \cite{c3, c4}. 

Much of the existing work has focused on the age-dependent effects of sleep on cognition. Della Monica et al. \cite{c5} highlighted significant associations between slow wave sleep, sleep continuity, rapid eye movement (REM) sleep, and cognition. Djonlagic et al. \cite{c6} identified 23 predictive sleep metrics, including REM duration and EEG features, as critical for cognitive performance, particularly processing speed. Sleep stage distribution across the night has also been linked to cognitive impairment in older adults \cite{c7}.

While many sleep–cognition studies have long focused on memory \cite{c8}, research linking sleep to executive function also has a substantial history \cite{c9}. More recent work, however, has begun zeroing in on specific components of sleep quality and their role in executive processes. For example, Sen et al. \cite{c10} found that low sleep efficiency and high fragmentation correlate with poorer executive function and heightened dementia risk, placing greater importance on sleep efficiency over mere duration. Abbas et al. \cite{c11} demonstrated the impact of poor sleep quality on response inhibition, underscoring the role of REM sleep in facilitating appropriate reactions. Meanwhile, Magnuson et al. \cite{c12} reported diminished inhibitory control and slower neural processing following sleep deprivation.

\begin{figure*}[!t]
\centerline{\includegraphics[width=17cm]{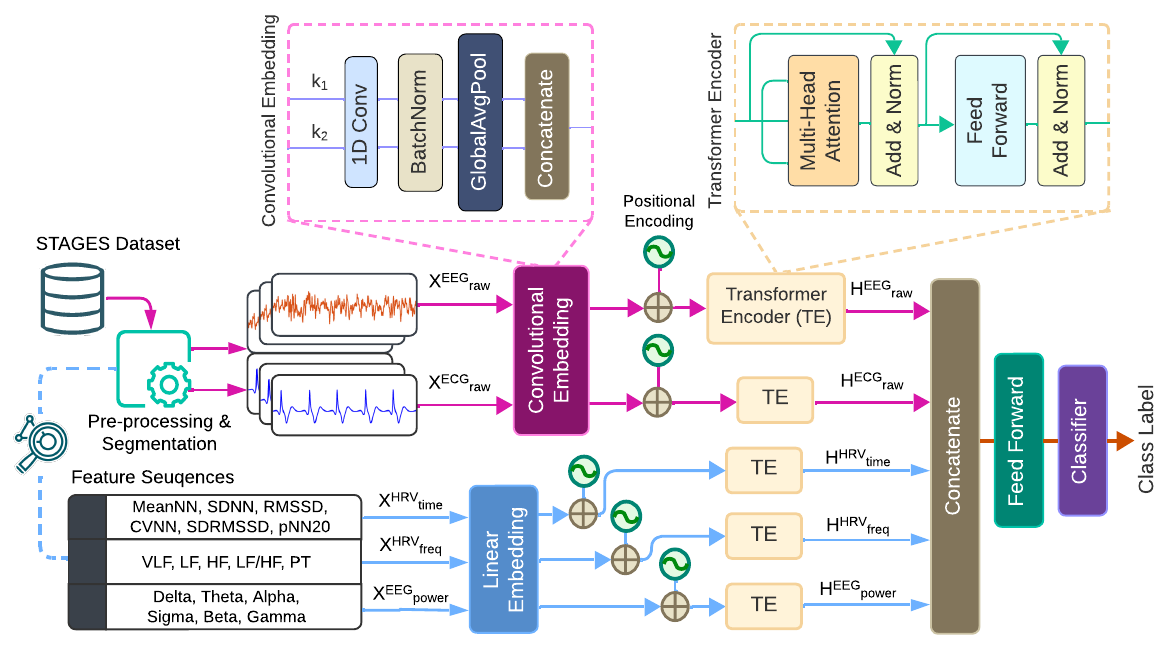}}
\caption{\centering Block diagram of the proposed CogPSGFormer architecture.}
\label{fig1}
\end{figure*}

Most studies have focused on age-related cognitive decline in older adults \cite{c6, c7, c10, c13}, limiting research on normal cognitive performance across age groups. Memory has been the primary focus, leaving other cognitive domains like conceptual reasoning and inhibition underexplored. Additionally, associations between cognitive function and sleep metrics (e.g., NREM/REM duration, total sleep time, sleep efficiency) have mostly been examined using statistical hypothesis testing or linear models \cite{c2, c9}. Some studies have attempted to incorporate physiological biomarkers, such as EEG power bands and heart rate variability (HRV) parameters \cite{c3}, which offer a richer representation of brain activity and autonomic function during sleep. However, these biomarkers have typically been treated as static, night-level features—calculated once per recording or study period—thereby limiting their ability to capture dynamic variations and temporal dependencies in sleep-cognition relationships.

While these studies have provided valuable insights, their reliance on simplified, aggregated sleep metrics may have overlooked complex, nonlinear patterns in sleep physiology. To address this limitation, advanced deep learning techniques and non-linear modeling approaches offer new opportunities to analyze raw, multi-channel sleep data and uncover intricate temporal patterns across the nocturnal sleep cycle.

To date, only one study has applied deep learning to cognition prediction from raw PSG signals, focusing on age-related cognitive impairment as an early marker of Alzheimer’s disease \cite{c14}. Their Transformer-based model, trained on 1,500 single-channel sleep EEG recordings, achieved 70.22\% accuracy, outperforming traditional methods. However, it was limited to cognitive decline in older adults, overlooking general cognitive performance across age groups. This underscores the need for a broader deep learning framework that integrates raw PSG data and extracted features to predict cognitive performance beyond clinical impairment, capturing multi-scale physiological patterns across diverse populations.

This study pioneers the development of a deep learning model to explore the relationship between sleep and general cognitive function, specifically executive functioning. We introduce CogPSGFormer, a novel hybrid CNN-Transformer model, to extract biomarkers from multi-channel PSG signals, leveraging one EEG and one ECG channel—selected based on established links between brain and heart activity and cognitive function—along with spectral and time-domain features. The model predicts the concept-level response index from the Penn Conditional Exclusion Test (PCET), emphasizing cognitive flexibility and response inhibition, which are critical for real-world decision-making in scenarios like late-night driving or early-morning activities.

Given the complexity of multi-channel, long-duration PSG data, this study tackles key challenges in building an effective prediction pipeline. We systematically analyze signal channels and temporal segments to identify the most informative components for cognitive performance. Various data processing and feature extraction strategies are compared, evaluating raw signal patterns against hand-crafted features. Additionally, we assess Transformer-based models versus recurrent neural networks (RNNs) for capturing long-term temporal dependencies in sleep. An ablation study contrasts CogPSGFormer with a vanilla Transformer and a single-scale CNN-Transformer to evaluate CNNs' role in feature extraction, considering both per-channel and cross-channel representations. By integrating learned biomarkers and hand-crafted features, our work reveals new associations between sleep and executive function.

\section{MATERIALS AND METHODS}

\subsection{Dataset Overview}

The dataset used in this paper is part of the Stanford Technology Analytics and Genomics in Sleep (STAGES) study \cite{c15}, a prospective cross-sectional, multi-site initiative. Data were collected across 20 sites from six centers in the USA and Canada, including Stanford University, Bogan Sleep Consulting, Geisinger Health, Mayo Clinic, MedSleep, and St. Luke's Hospital. Participants underwent in-lab sleep studies with at least 2 hours of PSG recording.

Exclusion criteria in the STAGES included individuals unable to complete required assessments, share blood samples, or pair a smartphone with an actigraph device. EEG and ECG were used as the primary input modalities based on extensive prior research linking EEG power bands and ECG HRV parameters \cite{c16} to cognitive function, as they provide crucial insights into brain activity and autonomic regulation, both essential for understanding cognition. For this research, one EEG channel (C3-M2) and one ECG channel (ECG-I) were utilized. Due to inconsistencies in data channels across and within clinics, individuals without these specific channels were excluded from the analysis. Additionally, participants with corrupted channels (e.g., signals containing mostly zero values) were removed from the dataset. The data were further filtered based on the availability of cognitive scores, resulting in a final cohort of 817 individuals (52\% male, 33\% with moderate to severe OSA (AHI>15)) with 2-11 hours of signal recordings. Table~\ref{tab1} summarizes the demographic characteristics of the study population. 

\begin{table}[t]
\caption{Demographic Overview of Study Participants (Mean ± SD or Count [\%]).}
\newcolumntype{C}{>{\centering\arraybackslash}X}
\renewcommand{\arraystretch}{1.2} 
\begin{tabularx}{250pt}{CC}
\toprule
\textbf{Characteristics} & \textbf{Value}\\
\midrule
Sample (no.)     & 817\\
Sex (male)		& 425 (52\%)\\
Age, years     & 44 ± 15.2\\
Body mass index, $kg/m^2$	& 29.5 ± 8.1\\
\bottomrule
\end{tabularx}
\label{tab1}
\end{table}

\subsection{PCET Concept Level Responses}

The PCET is a widely used neurocognitive task designed to assess executive function, particularly abstraction and cognitive flexibility \cite{c17}. In the STAGES study, this 6-minute test was administered either in-lab or off-site, typically the evening before PSG (prior to 9:30 PM) or the following morning (at least one hour after awakening). During PCET, participants are required to determine which of four objects does not belong to a group based on shifting abstract rules, such as color, shape, or texture. These rules change without warning, requiring participants to adapt their responses based on trial feedback.

The Concept Level Responses score is one of the main outputs which reflects the participant’s ability to correctly identify the abstract rule and consistently apply it across trials. Higher scores indicate better cognitive flexibility and rule learning, while lower scores suggest difficulties in abstraction or adaptability. The median score in the dataset was utilized as a threshold to transform the task into a binary classification, categorizing participants into two groups indicative of lower and higher cognitive performance.

\subsection{Signal Pre-Processing}

Among the available EEG channels, C3 was selected for its role in capturing frontocentral activity related to motor inhibition, response control, and executive function—key components of conceptual reasoning assessed by PCET \cite{c18}. The chosen channel was re-referenced to the mastoid channel, and the signals were filtered using a 4th-order Butterworth band-pass filter. The filter cutoff frequencies were set to 0.3–35 Hz for EEG signals and 0.5–35 Hz for ECG signals.
The YASA sleep analysis toolbox (version 0.6.5) was employed to identify significant EEG artifacts and generate a binary mask to exclude specific 5-second epochs. Subsequently, the ECG and EEG signals were segmented into 2-minute and 30-second epochs, respectively.

Feature extraction was performed as follows: 

The power spectral density of EEG signals was computed using the multitaper method. For each 30-second segment, power in the following frequency bands was calculated: Delta (0.5–4 Hz), Theta (4–8 Hz), Alpha (8–12 Hz), Sigma (12–16 Hz), Beta (16–30 Hz), and Gamma (30–35 Hz). For HRV parameters extracted from ECG signals, R-peaks were detected using the Neurokit library. Two-minute segments (with 50\% overlap) were used for time-domain HRV calculations to reliably capture short-term variability measures such as RMSSD and SDNN, which are optimized for shorter durations. Five-minute segments (with 50\% overlap) were chosen for frequency-domain HRV analysis to ensure accurate estimation of both low- and high-frequency components, as spectral analysis requires longer windows to achieve adequate frequency resolution and stability. 

To further ensure data quality, segments with an abnormal number of R-peaks (corresponding to heart rates outside the range of 40–100 bpm) were excluded to eliminate corrupted recordings. From the extracted HRV parameters, frequency-domain features included VLF, LF, HF, LF/HF, and TP, while time-domain features included MeanNN, SDNN, RMSSD, CVNN, SDRMSSD, and pNN20. It is important to note that each feature is extracted over time, forming a time series rather than a single aggregate value per night.

To maintain consistency in sequence length across the dataset, a uniform duration of 5 hours of data was selected for model training. Additionally, all raw and feature sequences were normalized using z-score standardization to achieve zero mean and unit standard deviation.

\subsection{CogPSGFormer}
\label{model}

To capture multi-modal polysomnographic features, we developed a novel architecture combining a multi-scale CNN with a Transformer-based framework called CogPSGFormer. The model is designed to process raw signals and extracted features, allowing it to leverage complementary information across diverse data modalities. We identified the optimal multi-modal, multi-scale architecture through an ablation study (refer to \ref{subsec: ablation} for more information). The architecture of the optimal configuration of CogPSGFormer with Multiple Scales, Separate Paths for Channels, and a Shared Path for Scales (MS-SC-SHS) is displayed in Fig.~\ref{fig1} and explained in more detail below.

\subsubsection{Input Representation}

The dataset consists of 5-hour recordings for both EEG and ECG, sampled at 70 Hz. The raw EEG signal is segmented into 30-second windows, while the raw ECG signal is segmented into 2-minute windows. The raw signals are represented as follows:
\begin{equation}
    X^{\text{EEG}}_{\text{raw}} \in \mathbb{R}^{N \times S_{30} \times T_{30}}
\end{equation}
\begin{equation}
    X^{\text{ECG}}_{\text{raw}} \in \mathbb{R}^{N \times S_{120} \times T_{120}}
\end{equation}
where $N$ is the number of individuals, $S_{30}$ and $S_{120}$ are the number of 30-second and 2-minute segments, respectively, and $T_{30}$ and $T_{120}$ represent the number of time steps per segment for EEG and ECG, respectively.

Extracted features include: 6 EEG power bands extracted per 30-second window, 6 time HRV parameters extracted per 5-minute window, and 5 frequency HRV parameters extracted per 2-minute window, represented as:
\begin{equation}
X^{\text{EEG}}{\text{power}} \in \mathbb{R}^{N \times S{30} \times 6}
\end{equation}
\begin{equation}
X^{\text{HRV}}{\text{time}} \in \mathbb{R}^{N \times S{300} \times 6}
\end{equation}
\begin{equation}
X^{\text{HRV}}{\text{freq}} \in \mathbb{R}^{N \times S{120} \times 5}
\end{equation}

\subsubsection{Model Architecture}

The model consists of two main processing streams. The raw signal stream processes EEG and ECG signals using convolutional embeddings, positional encoding, and Transformer encoder layers in two identical parallel paths. The feature stream handles EEG power bands, time-domain HRV parameters, and frequency-domain HRV parameters via linear embeddings, positional encoding, and Transformer encoder layers in three identical parallel paths.

Convolutional Embeddings: The convolutional embeddings are multi-scale, utilizing two different kernel sizes to capture both fine-grained and coarse-grained temporal patterns within each segment of EEG and ECG. These kernel sizes share the same convolutional weights. Instead of defining a CNN layer for each kernel size separately, the model defines a single convolutional layer with the largest kernel size and dynamically slices the weights to match the kernel size being used at each step. 

During the forward pass, the signal is processed segment by segment by the convolutional embedding. Given a raw input segment where $B$, $S$, and $T$ are respectively the batch size, the number of segments, and the segment length:
\begin{equation}
X_{\text{raw}} \in \mathbb{R}^{B \times S \times T}
\end{equation}
Each segment is passed through the shared CNN layer multiple times, each time using different-sized kernels by slicing the filter weights dynamically. A ReLU activation is applied immediately after the convolution, followed by batch normalization. The output for a given kernel size is:
\begin{equation}
C_{\text{conv}}^{(k)} = \text{BatchNorm}\left(\text{ReLU}\left(\text{Conv1D}(X_{\text{raw}}, W) + b \right)\right)
\end{equation}
where $W$ and $b$ are the shared convolutional weights and biases. Global average pooling then reduces each output to a fixed-dimensional representation:
\begin{equation}
C_{\text{pool}}^{(k)} = \text{GAP}(C_{\text{conv}}^{(k)})
\end{equation}
where \text{GAP} represents the global average pooling operation. The outputs from different kernel sizes are then concatenated to form a multi-scale feature representation for each segment:
\begin{equation}
C_{\text{multi}} = \left[C_{\text{pool}}^{(k_1)}, C_{\text{pool}}^{(k_2)}\right] \in \mathbb{R}^{S \times (2 \cdot d_{\text{conv}})}
\end{equation}
where $k_1$ and $k_2$ are the kernel sizes, $d_{\text{conv}}$ is the number of output channels per kernel, and 2.$d_{\text{conv}}$ is the dimension of the subsequent Transformer. A dropout layer is applied to improve generalization:
\begin{equation}
X_{\text{raw, trans}} = \text{Dropout}(C_{\text{multi}})
\end{equation}
In the feature stream, each feature sequence type is projected into the Transformer model dimension using a linear embedding layer. For each feature sequence:
\begin{equation}
X_{\text{feat, trans}} = W X_{\text{feat}} + b, \quad X_{\text{feat, trans}} \in \mathbb{R}^{S \times d_{\text{feat}}}
\end{equation}
where $d_{\text{feat}}$ is the dimension of the subsequent Transformer.

In both the raw and feature streams, the embeddings are further processed using fixed positional encoding to preserve sequential order information across the night:
\begin{equation}
X_{\text{raw, pos}} = X_{\text{raw, trans}} + PE(S)
\end{equation}
\begin{equation}
X_{\text{feat, pos}} = X_{\text{feat, trans}} + PE(S)
\end{equation}
These encoded representations then pass through Transformer encoder layers, which capture long-range dependencies and interrelationships across segments. The Transformer encoder operation is formally defined as:
\begin{equation}
Z = \mathrm{LN}\left(X_{\text{pos}} + \mathrm{Dropout}\left(\mathrm{MHA}(X_{\text{pos}})\right)\right)
\end{equation}
\begin{equation}
H = \mathrm{LN}\left(Z + \mathrm{Dropout}\left(\mathrm{FFN}(Z)\right)\right)
\end{equation}
where \text{MHA(.)} denotes the multi-head self-attention mechanism, \text{FFN(.)} is the feed-forward network, and \text{LN(.)} represents layer normalization.

Finally, the last time-step representation from each Transformer-encoded stream is extracted, 
\begin{equation}
H_{\text{final}} = H_{S}^{\text{EEG}} \oplus H_{S}^{\text{ECG}} \oplus H_{S}^{\text{time HRV}} \oplus H_{S}^{\text{freq HRV}} \oplus H_{S}^{\text{power}}
\end{equation}
where \( \oplus \) denotes concatenation across all modalities. The concatenated representation is finally passed through a fully connected layer and a binary classifier for prediction:
\begin{equation}
y = \sigma\left(W_{\text{cls}} \text{ReLU}\left(W_{\text{FFN}} H_{\text{final}} + b_{\text{FFN}}\right) + b_{\text{cls}}\right)
\end{equation}
where \( \sigma(.) \) represents the sigmoid activation function for binary classification.

\subsection{Training Setup}

Our model was trained and evaluated using a 10-fold cross-validation approach. In each fold, data from 90\% of the individuals were allocated for training and validation, with 15\% of this subset specifically used for hyperparameter optimization. Data splitting was performed in a subject-based manner to ensure that no overlapping samples or information from the same individual appeared in both the training and testing sets. Optimization was performed using the Adam optimizer to minimize binary cross-entropy loss. Model training and evaluation were performed on Compute Canada GPU clusters with high-performance NVIDIA GPUs and up to 100GB of memory, ensuring sufficient resources for the multi-scale CNN-transformer architecture and large sleep signal inputs.

After optimizing the hyperparameters during training, the following values were selected: a learning rate of $1e^{-4}$, 3 layers for the raw signal Transformer and 4 for the feature Transformer, each with 4 attention heads. The Transformer's feed-forward dimensions were 92 (raw) and 8 (features), while model dimensions were 64 (raw) and 8 (features). The dropout rate was 0.05, and the final feed-forward network dimension was 92. Table~\ref{tab2} summarizes the tunable hyperparameters and their search spaces. Fixed architectural components included a batch size of 4, 70 epochs, 1 CNN layer, and kernel sizes of 3 and 7. 

 \begin{table}[t]
\caption{Model Hyperparameters and Their Search Space}
\newcolumntype{C}{>{\centering\arraybackslash}X}
\renewcommand{\arraystretch}{1.2} 
\begin{tabularx}{250pt}{>{\raggedright\arraybackslash}p{3.4cm} p{3.1cm} p{2.5cm}} 
\toprule
\textbf{Hyperparameter}	& \textbf{Search space} & \hspace{-0.2em}\textbf{Selected}\\
\midrule
T layers (raw)	& \{2, 3, 4\} & 3\\
T layers (features)	& \{2, 3, 4\} & 4\\
Attention heads	& \{4, 8\} & 4\\
T FFN dimension (raw) & \{92, 128, 256, 512\} & 92\\
T FFN dimension (features) & \{8, 16, 32\} & 8\\
T model dimension (raw) & \{32, 64, 128\} & 64\\
T model dimension (features) & \{8, 16\} & 8\\
Dropout rate	& \{0.01, 0.05, 0.1, 0.2\} & 0.05\\
Classifier dimension & \{92, 128, 256\} & 92\\
Learning rate     & [1e-3, 1e-5] (log step 5) & 1e-4\\
\bottomrule
\end{tabularx}
\captionsetup{justification=centering}
\caption*{\small \textbf{Abbreviations:} T: Transformer, FFN: Feed-Forward Network}
\label{tab2}
\end{table}

\section{RESULTS}

\subsection{Comparative Analysis and Ablation Study}
\label{subsec: ablation}

To assess the effectiveness of CogPSGFormer and refine its design for predicting cognition from sleep data, we conducted a comprehensive analysis encompassing both comparisons with baseline models and an ablation study exploring different model configurations. This analysis aimed to answer key questions regarding input representation, model architecture, and feature extraction strategies.

First, to establish a performance benchmark, we compared CogPSGFormer against a vanilla Transformer and a stacked LSTM model. The vanilla Transformer was tested under three configurations: using raw signals, extracted features, and their combination. The results indicated that the best performance was achieved when using both raw and feature-based inputs, leading us to adopt this setting for all subsequent experiments. To determine the effectiveness of the Transformer architecture in capturing long-term dependencies, we compared it against a stacked LSTM model. The results confirmed that Transformers outperform LSTMs in modeling extended temporal relationships, reinforcing their suitability for this task.

Following these baseline comparisons, we conducted an ablation study to explore four different configurations of our hybrid CNN-Transformer (CogPSGFormer) to identify the optimal design.

\begin{enumerate}

\item CogPSGFormer single-scale: A single-kernel CNN for embedding raw EEG and ECG signals, using separate processing paths for each channel. 

The subsequent configurations (2 to 4) introduced multi-scale CNNs with two kernel sizes to enhance feature extraction across different temporal resolutions:

\item CogPSGFormer Multi-Scale, SHared path for Channels \& SHared Path for Scales (MS-SHC-SHS): EEG and ECG were concatenated into a single shared processing path, with one CNN embedding layer that dynamically adjusted filter weights to accommodate two different kernel sizes.
   
\item CogPSGFormer Multi-Scale, Separate Paths for Channels \& Shared Path for Scales (MS-SC-SHS): Separate paths were maintained for EEG and ECG, but a shared CNN embedding layer was used for multi-scale feature extraction within each channel. This configuration is explained in more detail in Section \ref{model}.

\item CogPSGFormer Multi-Scale, Separate Paths for Channels \& Separate Paths for Scales (MS-SC-SS): Both EEG and ECG had independent CNN processing paths, with separate embeddings for each of the two kernel sizes, ensuring maximum scale-specific representation.

\end{enumerate}

\begin{table}[t]
\caption{Evaluation of different architectures based on 10-fold cross-validation, SC-SS: separate paths for channels and scales, SC-SHS: separate paths for channels shared path for scales, SHC-SHS: shared path for channels and scales}
\newcolumntype{C}{>{\centering\arraybackslash}X}
\renewcommand{\arraystretch}{1.2} 
\begin{tabularx}{250pt}{>{\hspace{0.1em}}p{4.7cm} >{\hspace{0.5em}}p{1.5cm} >{\hspace{1em}}p{1.5cm}}
\toprule
\textbf{Architecture} & \hspace{-0.5em}\textbf{Accuracy} & \hspace{-0.8em}\textbf{F1-Score}\\
\midrule
Vanilla Transformer (only raw) & 64.04\% & 0.71\\
Vanilla Transformer (only features) & 71.10\% & 0.78\\
Vanilla Transformer (mixed) & 76.16\% & 0.84\\
Stacked LSTM (mixed) & 73.10\% & 0.80\\
CogPSGFormer single-scale (mixed) & 79.81\% & 0.88\\
CogPSGFormer MS-SHC-SHS (mixed) & 74.41\% & 0.81\\
CogPSGFormer MS-SC-SHS (mixed) & \textbf{80.30\%} & \textbf{0.89}\\
CogPSGFormer MS-SC-SS (mixed) & 78.94\% & 0.87\\
\bottomrule
\end{tabularx}
\label{tab3}
\end{table}

In all configurations of CogPSGFormer, time-domain HRV, frequency-domain HRV, and EEG powerband features were processed through independent paths, using a linear embedding layer similar to that of the vanilla Transformer. 

\subsection{Performance Evaluation} 

Table~\ref{tab3} presents the performance of various model architectures using 10-fold cross-validation. The results provide key insights into the comparative performance of RNN-based and Transformer-based models, as well as the impact of different architectural components within CogPSGFormer on classification accuracy.

To establish a performance baseline, we first compare CogPSGFormer against two widely used deep learning models: a vanilla Transformer and a stacked LSTM. The vanilla Transformer was evaluated with three different input configurations: raw signals, extracted features, and their combination. The results indicate that using only feature-based inputs (71.10\%) led to better performance than using raw signals alone (64.04\%), indicating that handcrafted features provide a more structured and informative representation for the model. However, combining both raw and feature-based inputs further improved accuracy to 76.16\%, confirming that deep learning can leverage complementary information from both representations. A similar Transformer model was also utilized in \cite{c14}, achieving 70.22\% accuracy in predicting cognitive decline in older adults using single-channel EEG combined with demographic data.

The stacked LSTM model underperformed compared to the Transformer-based models, reinforcing the idea that self-attention mechanisms are more effective than recurrent architectures in capturing long-range temporal dependencies. 

Beyond these baseline comparisons, we conduct an ablation study to analyze the contributions of different architectural components within CogPSGFormer. Incorporating CNNs for feature extraction before the Transformer led to a notable performance boost, with CogPSGFormer single-scale achieving 79.81\% accuracy. This 4.8\% improvement underscores the effectiveness of spatial feature extraction in enhancing temporal modeling and capturing more meaningful representations from sleep signals. 

The multi-scale CogPSGFormer models provide additional insights into the importance of proper scale integration. Sharing both scale and channel pathways in MS-SHC-SHS resulted in 74.41\% prediction accuracy, likely due to excessive feature compression across signal channels leading to a loss of critical information. The MS-SC-SS model (78.94\%) outperformed the vanilla Transformer and LSTM. However, its accuracy was approximately 1\% lower than the single-scale model. This can be attributed to its significantly higher parameter count and increased model complexity relative to the dataset size, which may have led to overfitting. In contrast, the MS-SC-SHS model, which maintains separate paths for EEG and ECG channels but shares convolutional paths for scales, achieves the highest accuracy (80.30\%). This result implies that sharing information across scales while preserving channel distinctions enhances feature fusion for cognitive prediction. 

These findings confirm the effectiveness of CNN-based feature extraction and the advantages of the Transformer architecture in improving classification performance for extended PSG data. Importantly, they emphasize that while multi-scale processing can be beneficial, careful design choices regarding feature fusion and channel separation are crucial to maximizing predictive accuracy.

\section{Discussion} 

While the link between sleep architecture and cognition has been widely studied, how sleep microstructure shapes cognitive performance across domains remains unclear. More critically, whether deep learning can predict these relationships from overnight physiological sleep data is still an open question. Developing an effective pipeline for this under-explored task requires selecting the right data modalities and learning meaningful representations. This study is among the first to leverage deep learning for PSG-based cognition prediction, with a focus on executive function.

CogPSGFormer, a hybrid CNN-Transformer, automates biomarker extraction from multi-scale PSG signals. Multi-scale convolution captures fine-grained sleep features like spindles, while Transformer encoders recognize long-term sleep patterns. This dual capability provides a comprehensive representation of sleep dynamics relevant to cognition.

A key question in physiological signal analysis is whether deep learning can extract task-relevant patterns from raw signals or if feature engineering remains essential. To investigate this, we systematically evaluated three configurations of our baseline Transformer: raw data only, engineered features only, and a combination of both. Results on the STAGES dataset (Table~\ref{tab3}) show that relying solely on raw signals was suboptimal, achieving an accuracy of 64.04\%, while engineered features provided a more structured representation for the model (71.10\%). Combining both further improved performance (76.16\%), demonstrating that deep learning benefits from the complementary nature of raw and engineered inputs in cognition prediction.

We tested the hypothesis that Transformer models better capture long-term dependencies in sleep signals than traditional sequential models like LSTM. Results support this: while the stacked LSTM reached 73.10\% accuracy on mixed inputs, Transformer variants ranged from 74.41\% to 80.30\%. This gap highlights the advantage of self-attention in modeling extended temporal relationships, whereas LSTMs may be limited by vanishing gradients and memory constraints. 

Although pure Transformer models have demonstrated strong capabilities in extracting meaningful representations for vision tasks without relying on CNNs, prior research across various domains suggests that integrating CNNs and Transformers within a hybrid framework can potentially enhance representation learning and prediction. We explored whether this synergy holds for extracting sleep biomarkers correlated with cognitive function from PSG signals. Given that sleep events occur at different temporal scales (e.g., 0.5–3 second spindles, 0.5–1.5 second K-complexes), we investigated whether incorporating multiple convolutional kernel sizes could enhance feature extraction before feeding them into the Transformer encoder layers. 

We compared multiple CogPSGFormer configurations to identify optimal strategies for feature extraction and fusion. Adding CNN-based feature extraction improved accuracy by over 3 percentage points (76.16\% to 79.81\%). However, the impact of multi-scale convolution depended on the feature fusion strategy. The best-performing model—using separate EEG and ECG paths with shared convolutional layers across scales—outperformed the single-scale version by 1.5 percentage points (80.30\% vs 79.81\%). In contrast, fully merging features across channels and scales reduced performance to 74.41\%, suggesting that excessive compression hinders the model’s ability to capture fine-grained biomarkers. Larger, more complex models also slightly overfit, reducing accuracy by 1 point. These results highlight that while CNNs enhance Transformer-based models, performance depends heavily on how multi-scale features are structured and fused.

The only closely related study, by Song et al.~\cite{c14}, used a transformer encoder to predict cognitive impairment from single-channel sleep EEG with sleep stage labels and demographic data. While their work focused on CASI score prediction from EEG alone, our CogPSGFormer extends this by incorporating multimodal signals (EEG and ECG) and combining convolutional and Transformer layers to jointly model local and global dependencies. Although performance is not directly comparable due to different prediction targets and cohorts, our approach builds on this foundation and underscores the benefits of multimodal modeling and structured feature extraction for sleep-based cognitive prediction.

This study has several limitations. To balance efficiency, memory constraints, and overfitting risks, we designed a streamlined architecture with a limited number of trainable parameters. Our analysis focused on one EEG/ECG channel and selected spectral and HRV features, though broader multi-modal representations may offer additional insights. Expanding model complexity and feature diversity in future investigations could refine representation learning.

Deep learning models are often perceived as black boxes, posing interpretability challenges. A deeper analysis of Transformer attention patterns and visualization techniques could help clarify how sleep features contribute to cognition prediction, enhancing alignment with domain expertise. Transformers offer a more structured and interpretable design than LSTMs, which compress sequential information into hidden states, making input influence hard to trace. In contrast, transformers provide explicit attention scores, highlighting which parts of the sleep signal the model focuses on. While not fully causal, attention enables more transparent post-hoc analysis and feature attribution. Both models capture temporal patterns, but transformers provide clearer explainability.

Cognitive function is influenced by demographic factors such as age and education, and therefore, integrating this information could enhance prediction performance. Incorporating sleep stage annotations could also enhance learning \cite{c14}, which was not considered here due to inconsistencies in the STAGES dataset annotation and the infeasibility of contacting data managers. Additionally, while this study used binary classification, future work could explore regression-based approaches to predict cognitive scores more precisely. 

This study bridges sleep science and deep learning by showcasing the potential of hybrid CNN-Transformer models for cognition prediction from PSG signals. Through systematic evaluation, we demonstrate how multi-scale convolution and attention mechanisms capture critical physiological biomarkers. Beyond sleep research, this work supports applications in personalized sleep optimization, early cognitive decline detection, and understanding sleep microstructure's role in executive function. Future directions include demographic-aware modeling, additional physiological inputs, and enhanced interpretability for real-world deployment.

\section{CONCLUSIONS}

This study investigated the link between sleep signals and executive function using CogPSGFormer, a multi-scale Convolutional-Transformer model for PSG analysis. By integrating single-channel EEG and ECG with features like EEG power bands and HRV metrics, the model captured complementary physiological patterns. Evaluation on the STAGES dataset showed strong performance, achieving 80.3\% accuracy in classifying cognitive levels based on PCET scores. These results underscore the potential of deep learning to extract sleep-based biomarkers for cognitive assessment. Future work could extend this work by incorporating additional channels and exploring regression tasks to predict continuous cognitive scores for personalized health monitoring.

\addtolength{\textheight}{-12cm}   





\section*{ACKNOWLEDGMENT}

MdZ is currently an employee of Oura Health Oy. His contributions to this work, as well as the statements and opinions expressed, are solely his own and do not represent the official views of Oura Health Oy. MdZ has also received research funding unrelated to this work from Noctrix Health Inc. and Verily Life Sciences LLC.


\begin{thebibliography}{99}

\bibitem{c1} M. C. Deak and R. Stickgold, ``Sleep and cognition,'' Wiley Interdisciplinary Reviews: Cognitive Science, vol. 1, no. 4, pp. 491--500, 2010.

\bibitem{c2} N. C. Lima et al., ``Impairment of executive functions due to sleep alterations: An integrative review on the use of P300,'' Frontiers in Neuroscience, vol. 16, p. 906492, 2022.

\bibitem{c3} S. A. Immanuel et al., ``Symbolic dynamics of sleep heart rate variability is associated with cognitive decline in older men,'' in Proc. 45th Annu. Int. Conf. IEEE Eng. Med. Biol. Soc. (EMBC), 2023, pp. TBD.

\bibitem{c4} J. C. Lo et al., ``Cognitive performance, sleepiness, and mood in partially sleep deprived adolescents: the need for sleep study,'' Sleep, vol. 39, no. 3, pp. 687--698, 2016.

\bibitem{c5} C. Della Monica et al., ``Rapid eye movement sleep, sleep continuity, and slow wave sleep as predictors of cognition, mood, and subjective sleep quality in healthy men and women, aged 20--84 years,'' Frontiers in Psychiatry, vol. 9, p. 255, 2018.

\bibitem{c6} I. Djonlagic et al., ``Macro and micro sleep architecture and cognitive performance in older adults,'' Nature Human Behaviour, vol. 5, no. 1, pp. 123--145, 2021.

\bibitem{c7} ``Relationships between sleep stages and changes in cognitive function in older men: the MrOS Sleep Study,'' Sleep, vol. 38, no. 3, pp. 411--421, 2015.

\bibitem{c8} K. A. Paller et al., ``Memory and sleep: How sleep cognition can change the waking mind for the better,'' Annual Review of Psychology, vol. 72, pp. 123--150, 2021.

\bibitem{c9} J. Horne, ``Overnight sleep loss and 'executive' decision making—subtle findings,'' Sleep, vol. 36, no. 6, pp. 823--824, 2013.

\bibitem{c10} A. Sen and X. Y. Tai, ``Sleep duration and executive function in adults,'' Current Neurology and Neuroscience Reports, 2023.

\bibitem{c11} N. H. Abbas et al., ``The effects of sleep quality on response inhibition,'' Young Anthropology, vol. 2, pp. 10--16, 2020.

\bibitem{c12} J. R. Magnuson et al., ``Neural effects of sleep deprivation on inhibitory control and emotion processing,'' Behavioural Brain Research, vol. 426, p. 113845, 2022.

\bibitem{c13} M. Casagrande et al., ``Sleep quality and aging: a systematic review on healthy older people, mild cognitive impairment and Alzheimer’s disease,'' International Journal of Environmental Research and Public Health, vol. 19, no. 14, p. 8457, 2022.

\bibitem{c14} T.-A. Song et al., ``A transformer model for predicting cognitive impairment from sleep,'' bioRxiv, 2022.

\bibitem{c15} G.-Q. Zhang et al., ``The National Sleep Research Resource: towards a sleep data commons,'' Journal of the American Medical Informatics Association, vol. 25, no. 10, pp. 1351--1358, 2018.

\bibitem{c16} Alba, Guzmán, et al. "The relationship between heart rate variability and electroencephalography functional connectivity variability is associated with cognitive flexibility." Frontiers in Human Neuroscience 13 (2019): 428262.

\bibitem{c17} M. M. Kurtz et al., ``The Penn Conditional Exclusion Test: a new measure of executive-function with alternate forms for repeat administration,'' Archives of Clinical Neuropsychology, vol. 19, no. 2, pp. 191--201, 2004.

\bibitem{c18} C. de Klerk et al., ``An EEG study on the somatotopic organisation of sensorimotor cortex activation during action execution and observation in infancy,'' Developmental Cognitive Neuroscience, vol. 15, pp. 1--10, 2015.

\end{thebibliography}
\end{document}